\documentclass[12pt]{article}
\usepackage[round]{natbib}
\usepackage[T1]{fontenc}
\usepackage{amsthm}
\usepackage{graphicx}
\usepackage{hyperref}
\usepackage{setspace}
\usepackage{rotating}
\usepackage{enumitem}
\usepackage{grffile}
\usepackage{longtable}
\usepackage{placeins}
\usepackage{xcolor}
\usepackage[framemethod=TikZ]{mdframed}
\usepackage{dcolumn}
\usepackage{float}
\usepackage{tikz}
\usetikzlibrary{arrows.meta, positioning}
\usepackage{caption}

\usepackage{tabularray}
\usepackage{float}
\usepackage{graphicx}
\usepackage{codehigh}
\usepackage[normalem]{ulem}
\UseTblrLibrary{booktabs}
\UseTblrLibrary{siunitx}

\NewTableCommand{\tinytableDefineColor}[3]{\definecolor{#1}{#2}{#3}}

\usepackage{amsthm, amssymb, amsmath, pdfpages, url, relsize, fullpage, ccaption, subfig}

\newcommand{\SampleN}{2144}

\newcommand{\ControlChangeMean}{-1.20}
\newcommand{\ControlChangeSE}{0.06}
\newcommand{\ControlChangePctSD}{83.6}

\newcommand{\HeavyChangeMean}{-0.86}
\newcommand{\HeavyChangeSE}{0.06}
\newcommand{\LightChangeMean}{-0.93}
\newcommand{\LightChangeSE}{0.06}

\newcommand{\AttenuationHeavyPct}{28}
\newcommand{\AttenuationHeavyCILow}{16}
\newcommand{\AttenuationHeavyCIHigh}{40}

\newcommand{\AttenuationLightPct}{23}
\newcommand{\AttenuationLightCILow}{9}
\newcommand{\AttenuationLightCIHigh}{35}

\newcommand{\ReversalPctControl}{34.4}
\newcommand{\ReversalPctHeavy}{22.1}

\newcommand{\CATEPartyDiffPval}{0.40}

\newcommand{\PersuasivenessD}{-0.31}
\newcommand{\PersuasivenessCILow}{-0.42}
\newcommand{\PersuasivenessCIHigh}{-0.21}
\newcommand{\PersuasivenessPval}{$<$0.001}

\newcommand{\UseAgainD}{-0.20}
\newcommand{\UseAgainCILow}{-0.31}
\newcommand{\UseAgainCIHigh}{-0.10}
\newcommand{\UseAgainPval}{$<$0.001}

\newcommand{\AITrustD}{-0.10}
\newcommand{\AITrustCILow}{-0.17}
\newcommand{\AITrustCIHigh}{-0.04}
\newcommand{\AITrustPval}{0.002}

\newcommand{\AIPolicyD}{-0.24}
\newcommand{\AIPolicyCILow}{-0.35}
\newcommand{\AIPolicyCIHigh}{-0.13}
\newcommand{\AIPolicyPval}{$<$0.001}

\title{Perceived Political Bias in LLMs \\ Reduces Persuasive Abilities}
\author{Matthew DiGiuseppe\footnote{Corresponding Author: mdigiuseppe@gmail.com } \\ \& \\ Joshua Robison \\ \\ Institute of Political Science  \\ Leiden University}
\date{} 

\begin{document}
\maketitle
\thispagestyle{empty}

\begin{abstract}

\noindent Conversational AI has been proposed as a scalable way to correct public misconceptions and spread misinformation. Yet its effectiveness may depend on perceptions of its political neutrality. As LLMs enter partisan conflict, elites increasingly portray them as ideologically aligned. We test whether these credibility attacks reduce LLM-based persuasion. In a preregistered U.S. survey experiment (N=\SampleN), participants completed a three-round conversation with ChatGPT about a personally held economic policy misconception. Compared to a neutral control, a short message indicating that the LLM was biased against the respondent's party attenuated persuasion by \AttenuationHeavyPct\%. Transcript analysis indicates that the warnings alter the interaction: respondents push back more and engage less receptively. These findings suggest that the persuasive impact of conversational AI is politically contingent, constrained by perceptions of partisan alignment.

\end{abstract}

  \noindent The preregistration of this study can be found {\href{https://osf.io/nr7b2}{here.} This work is funded by European Research Council Starting Grant \#852334. The authors would like to thank Tobias Heinrich for early comments and advice and participants at the participants of the ``Politics of AI'' Workshop hosted by the Institut Barcelona d'Estudis Internacionals in September 2025. 

\clearpage

\noindent Large Language Models (LLMs) have demonstrated a remarkable capacity to persuade and change beliefs, including about political matters, with the potential to scale \citep{Costello2024,hackenburg2024evaluating,rogiers2024persuasion,Hackenburg2025,argyle2025testing,lin2025persuading,bai2025llm,offer2026deep,karinshak2023working,goel2025artificial,boissin2025dialogues}.\footnote{This study was reviewed and approved by the Ethics Review Committee Social Sciences of Leiden University's Faculty of Social Sciences.} This capacity is a double-edged sword \citep{summerfield2025impact}. LLMs can serve a fact-checking role, aligning the public with expert consensus on important issues such as climate change \citep{czarnek2025addressing}, and can reduce harmful conspiracy beliefs \citep{Costello2024}. Conversely, malicious actors can use them to spread politically expedient falsehoods, change voting intentions, and promote conspiracy beliefs \citep{lin2025persuading,costello2026large}.

Existing research suggests that LLMs will meaningfully alter the political landscape if, and when, they are deployed at scale. However, prior LLM persuasion experiments have been conducted in a context wherein LLMs have been relatively weakly connected to existing patterns of political conflict. Indeed, most respondents in our control group do not yet have strong opinions on the political biases of LLMs (see Fig. \ref{fig:3}). Current trends suggest that this period of low politicization is coming to an end as elite actors increasingly question the political neutrality of popular LLMs. US President Donald Trump has frequently claimed that Artificial Intelligence is ``woke'' (i.e., socially progressive) and issued an executive order entitled ``Preventing Woke AI in the Federal Government.''\footnote{Rozen C. US to mandate AI vendors measure political bias for federal sales. Reuters. December 11, 2025. The White House. Fact Sheet: President Donald J. Trump Prevents Woke AI in the Federal Government. July 23, 2025.} Trump ally and founder of Xai, Elon Musk, has likewise proclaimed that the left-wing bias in AI stems from ``woke'' training data.\footnote{ABC News. Elon Musk slams AI ‘bias’ and calls for ‘TruthGPT.’ Experts question his neutrality. Apr 19, 2023.} These claims find purchase in nascent research showing that LLM content tends to have  a left-wing slant in the US when discussing political issues \citep{rozado2024political,motoki2024more,westwood2025measuring}. Prominent Democrats, meanwhile, have questioned the neutrality of LLMs by pointing out that their output may reflect existing group-based biases in human-generated data.\footnote{Ocasio-Cortez A. (2019, Jan 27). “When you don’t address human bias, that bias gets automated …” X (formerly Twitter). https://x.com/aoc/status/1089560176787632129.} For instance, some US-based research demonstrates gender and racial biases in LLM output when summarizing political news \citep{fang2024bias}. Finally, growing campaign donations by AI companies, and their executives, further enmesh these tools in partisan conflict and have even inspired a nascent ``QuitGPT" movement among left-wing organizers \footnote{New York Times. The Richest 2026 Players: A.I., Crypto, Pro-Israel Groups and Trump. https://www.nytimes.com/2026/02/01/us/politics/ai-crypto-trump-super-pac-israel-2026-midterms.html; MIT Technology Review. A ``QuitGPT`` campaign is urging people to cancel their ChatGPT subscriptions. https://www.technologyreview.com/2026/02/10/1132577/a-quitgpt-campaign-is-urging-people-to-cancel-chatgpt-subscriptions/}. LLMs are becoming entrenched within the familiar pattern of partisan politics. 

The growing politicization of LLMs is critical to investigate because decades of research show that beliefs about the source of a persuasive message, and the extent of political conflict in the persuasion context, substantially influence whether persuasion occurs \citep{hovland1953, Zaller1992, lupia1998, druckman2001, Pornpitakpan2004, druckman2013, robison2022}. However, existing work on LLM persuasion has not yet adequately examined whether its influence is constrained by claims of political bias. One study examines this question by considering whether LLM persuasion is moderated by beliefs about the neutrality of LLMs \citep{goel2025artificial}. However, the study measures neutrality beliefs rather than manipulates them, leaving open the possibility of confounding causal processes. Ultimately, it remains to be seen whether the persuasive ability of LLMs can withstand insertion into a polarized political environment. This raises an important question and the opportunity to answer it: \emph{Can the persuasive abilities of LLMs withstand attacks to their perceived political neutrality?}

Answering this question allows us to better understand the boundary conditions of persuasion in the polarized environments where most political communication takes place. As such, we can update our expectations of how likely LLMs may shift the broader information environment and consequently AI's impact on democracy \citep{summerfield2025impact,simonMisinformationReloadedFears2023}. Next, by exploring the role of perceived bias in LLMs, we can better understand the levers of LLM persuasion (e.g., \cite{Hackenburg2025}). If persuasion drops due to concerns of political bias, then perceived neutrality may be one of the reasons LLM conversations are, currently, persuasive beyond personalization or arguments and strategic information overload \citep{Hackenburg2025,lin2025persuading}. This has clear importance for policymakers and developers. If concerns about bias undermine the ability of LLMs to correct misconceptions, then the epistemic benefits of the technology may be unevenly distributed across the political spectrum. For developers, understanding concerns of bias allows for more informed alignment strategies and the importance of managing user perceptions.

We test these questions in a preregistered U.S. survey experiment that randomizes messages about potential LLM biases prior to a short ChatGPT conversation. Our strongest message reduces persuasion by \AttenuationHeavyPct\% (95\% CI: \AttenuationHeavyCILow\%--\AttenuationHeavyCIHigh\%) relative to the control group.

\section*{Results}

In December 2025 and January 2026, we conducted a preregistered four-arm, between-subjects survey experiment (N=\SampleN) in the United States in which respondents engaged in a three-round conversation with a non-reasoning version of ChatGPT (4.1) about one of six economic policy topics for which their views were misaligned with the consensus among orthodox academic economists (see \nameref{MnM}). \citep{hackenburg2025scaling} indicates that newer, reasoning-based models offer limited additional returns. Still, the model we chose was a state-of-the-art non-reasoning model when we fielded the survey.} ChatGPT was prompted only with the question details, the respondent's answer, and instructions to persuade the respondent to adopt the consensus view while remaining truthful.\footnote{See SI for the full prompt.} Respondents were randomly assigned to one of four groups. One quarter of respondents saw a blank control screen prior to the conversation. A second group received information noting a non-directional bias that AI labs attempt to correct (Undirected Bias condition). The final two groups received information warning that ChatGPT was trained on data skewed toward out-partisan sources (Out-Party Bias: Light and Strong conditions). The stronger version of this treatment also included information about the potential political affiliations of ChatGPT founder Sam Altman and a picture of Altman with a politician from the respondent's opposing party (President Trump/House Speaker Pelosi), further implying an out-group bias. Respondents conversed with the LLM after receiving this information, after which we again asked them about their initial economic misconception and several questions about the conversation and future use of AI.

\begin{figure}
    \centering
    \includegraphics[width=.5\linewidth]{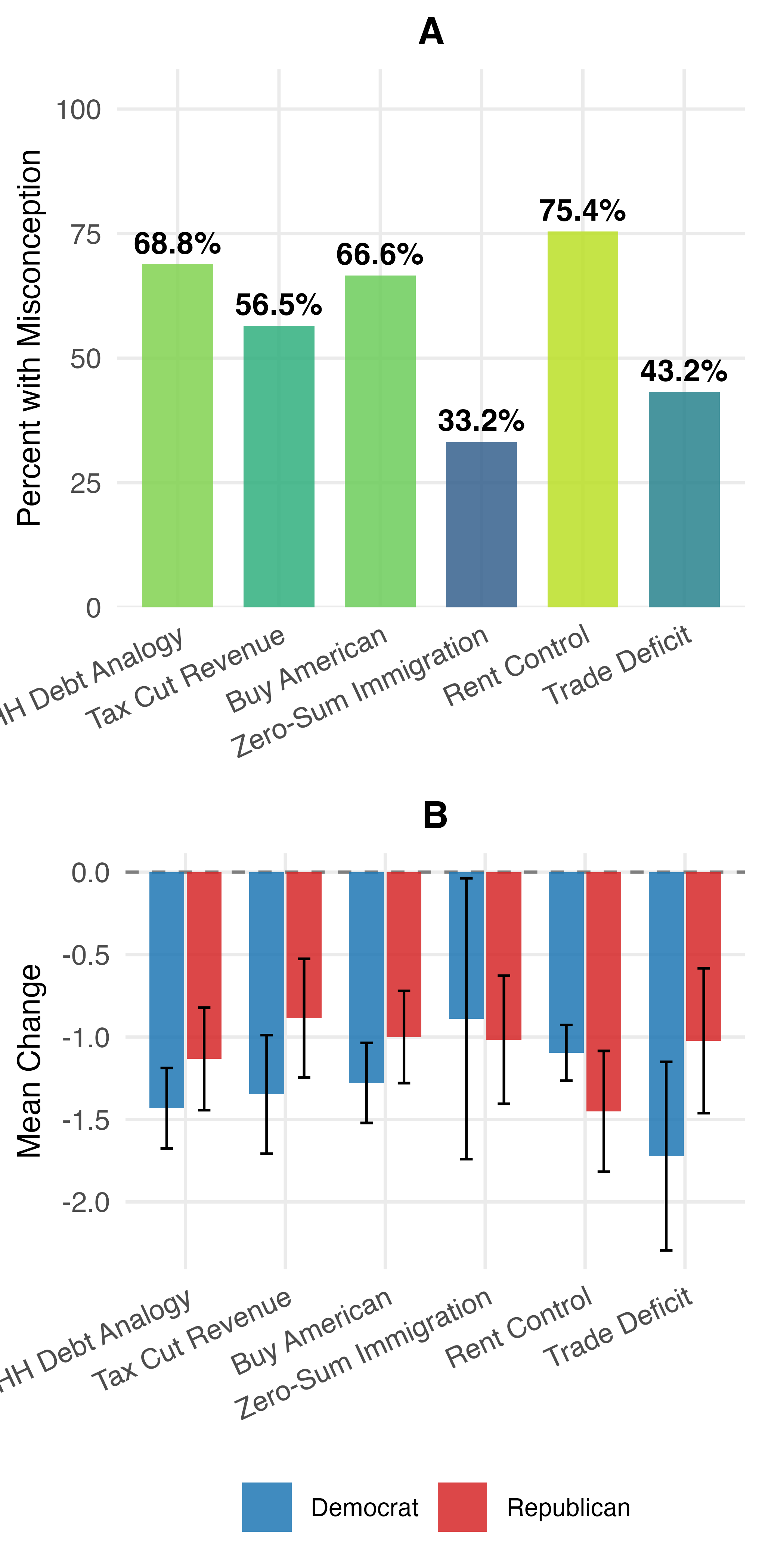}
    \caption{A. Percent of respondents in the final sample reporting at least moderate agreement with each economic misconception. B. Mean change (and SE) in agreement with the misconception on a 0--4 scale in the control arm, by party affiliation (including those who were forced to indicate closeness to one of the two parties).}
    \label{fig:1}
\end{figure}

In our initial respondent pool (N=2{,}414), 93\% of respondents held at least one of the six economic misconceptions in our analysis (the household--government budget analogy, the purported benefits of rent control, the zero-sum view of immigration and domestic jobs, the belief that trade deficits shrink national wealth, the belief that tax cuts pay for themselves, and the belief that ``Buy American'' policies increase national wealth). To reduce acquiescence bias, we first asked respondents to choose which of two statements they agreed with---the misconception or a statement reflecting the consensus view among academic economists (see \nameref{MnM}). We then measured agreement with the chosen statement on a five-point scale (``none at all'' to ``a great deal''). We code respondents as holding a misconception if they reported at least moderate agreement with the misconception statement. If a respondent held more than one misconception, we randomly assigned the conversation topic from that set. Fig.~\ref{fig:1}A shows the distribution of misconceptions in the final sample.

Fig.~\ref{fig:1}B shows the distribution of misconceptions by topic and respondents' partisanship in the control arm. Consistent with prior work, the LLM persuaded respondents across issue areas and partisan groups. On average, the raw pre--post change was \ControlChangeMean\ (SE=\ControlChangeSE) on a 0--4 scale, or \ControlChangePctSD\% of a standard deviation.

\begin{figure}
    \centering
    \includegraphics[width=1\linewidth]{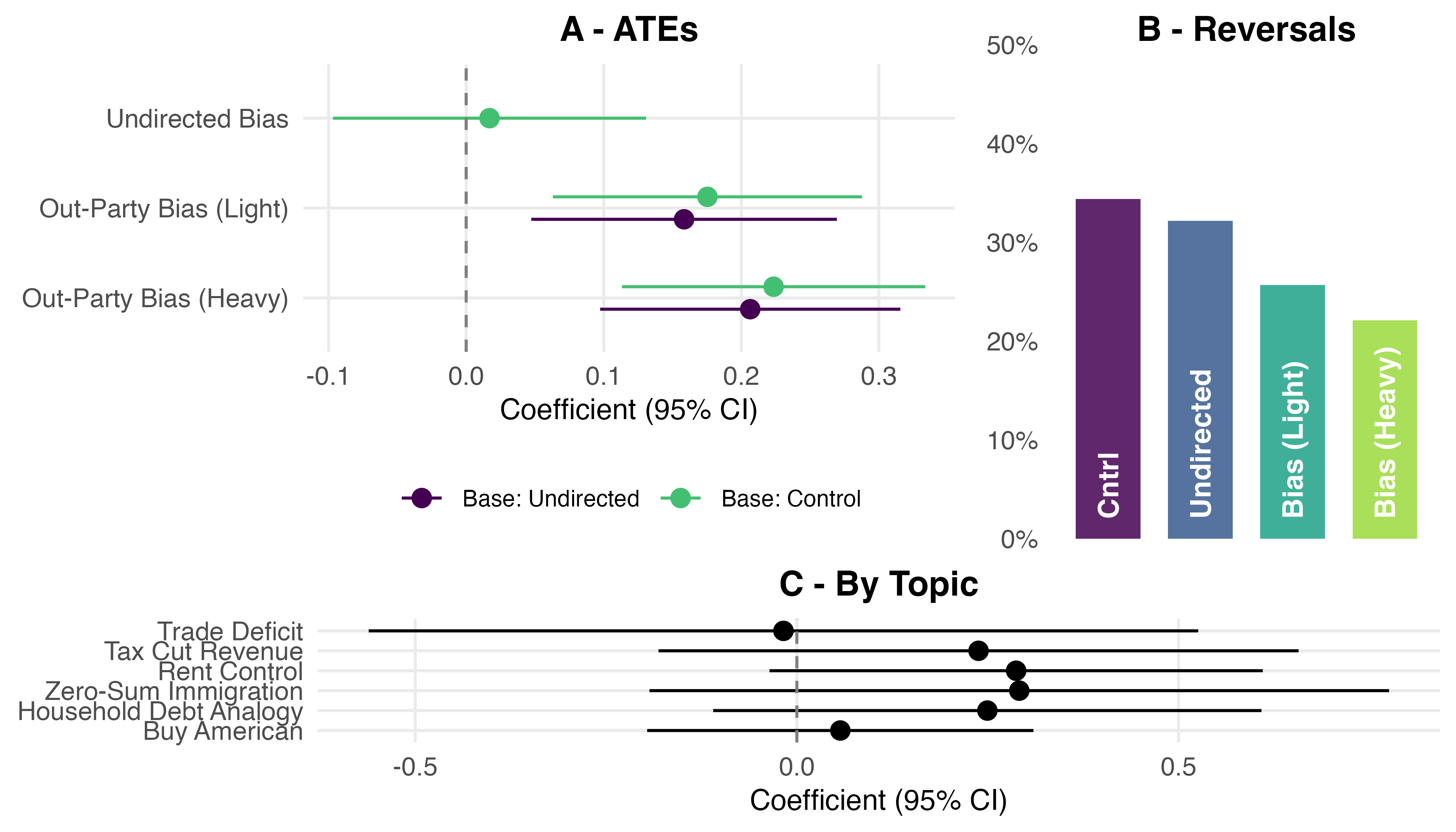}
    \caption{\textbf{A:} Standardized treatment effects (N=\SampleN) for each arm relative to the no-information condition or the non-directional bias condition. Point estimates and 95\% confidence intervals are from OLS models of post-treatment misconception agreement, including topic fixed effects and pretreatment agreement. \textbf{B:} Percent of respondents in each condition who moved from at least moderate agreement with a misconception to either no agreement with the misconception or agreement with the consensus position. \textbf{C:} Standardized coefficients and 95\% confidence intervals for topic-by-treatment interactions from an OLS model including the heavy treatment arm, the control arm, topic fixed effects, and their interactions, as well as pretreatment agreement. See \ref{si:mainfx} for the raw coefficients.}
    \label{fig:2}
\end{figure}

\subsection*{Main Findings}

Fig.~\ref{fig:2}A presents the standardized average treatment effects (ATEs) and 95\% confidence intervals for each condition relative to the no-information control or the non-directional bias arm. These estimates come from two OLS models of post-treatment misconception agreement that include treatment indicators, topic fixed effects, and pretreatment agreement.\footnote{We preregistered that we would also include additional covariates selected by a LASSO estimator to increase precision. This provided only marginal efficiency gains, so we omit covariates from the reported models.} Positive values indicate greater agreement with the misconception and, therefore, less persuasion. Our preregistered hypothesis was that the heavy treatment arm would have a positive and statistically significant effect relative to the control. The results support this hypothesis. The light treatment is similar in magnitude, and both treatments differ significantly from the non-directional bias arm.

In the control group, the mean pre--post change in misconception 
agreement was \ControlChangeMean\ (SE=\ControlChangeSE) on a 0--4 
scale. The corresponding change was \HeavyChangeMean\ 
(SE=\HeavyChangeSE) in the heavy treatment group and 
\LightChangeMean\ (SE=\LightChangeSE) in the light treatment group. 
This corresponds to a \AttenuationHeavyPct\% reduction in persuasion 
in the heavy condition (\AttenuationHeavyCILow--\AttenuationHeavyCIHigh\%) 
and a \AttenuationLightPct\% reduction in the light condition 
(\AttenuationLightCILow--\AttenuationLightCIHigh\%).\footnote{Because 
the attenuation percentage is a ratio of means, we computed 95\% 
confidence intervals using a nonparametric bootstrap with 2{,}000 
resamples, stratified by treatment condition.}

Fig.~\ref{fig:2}C presents results from an OLS model that restricts the sample to the control and heavy treatment groups and includes treatment-by-topic interactions. The out-party bias cue has a strong positive (i.e., persuasion-undermining) effect in four topic areas, whereas two topics have point estimates near zero. Given the small topic-specific sample sizes, none of the interaction estimates are statistically significant. Overall, this pattern suggests that the pooled effect is not driven by a single topic, although further interpretation risks conflating topic characteristics with the characteristics of respondents assigned to each topic.

Fig.~\ref{fig:2}B shows the percentage of respondents in each condition who fully reversed their views---moving from at least moderate agreement with the misconception to either ``no agreement at all'' or agreement with the opposing statement. In the control group, \ReversalPctControl\% of respondents reversed their views, compared to \ReversalPctHeavy\% in the heavy treatment group. This indicates that, beyond marginal decreases in agreement with the misconception, the bias messages also strongly influence full opinion reversals.

\subsection*{Heterogeneous Effects}

\begin{figure}[htbp]
\centering
\begin{minipage}[c]{0.55\linewidth}
  \centering
  \includegraphics[width=\linewidth]{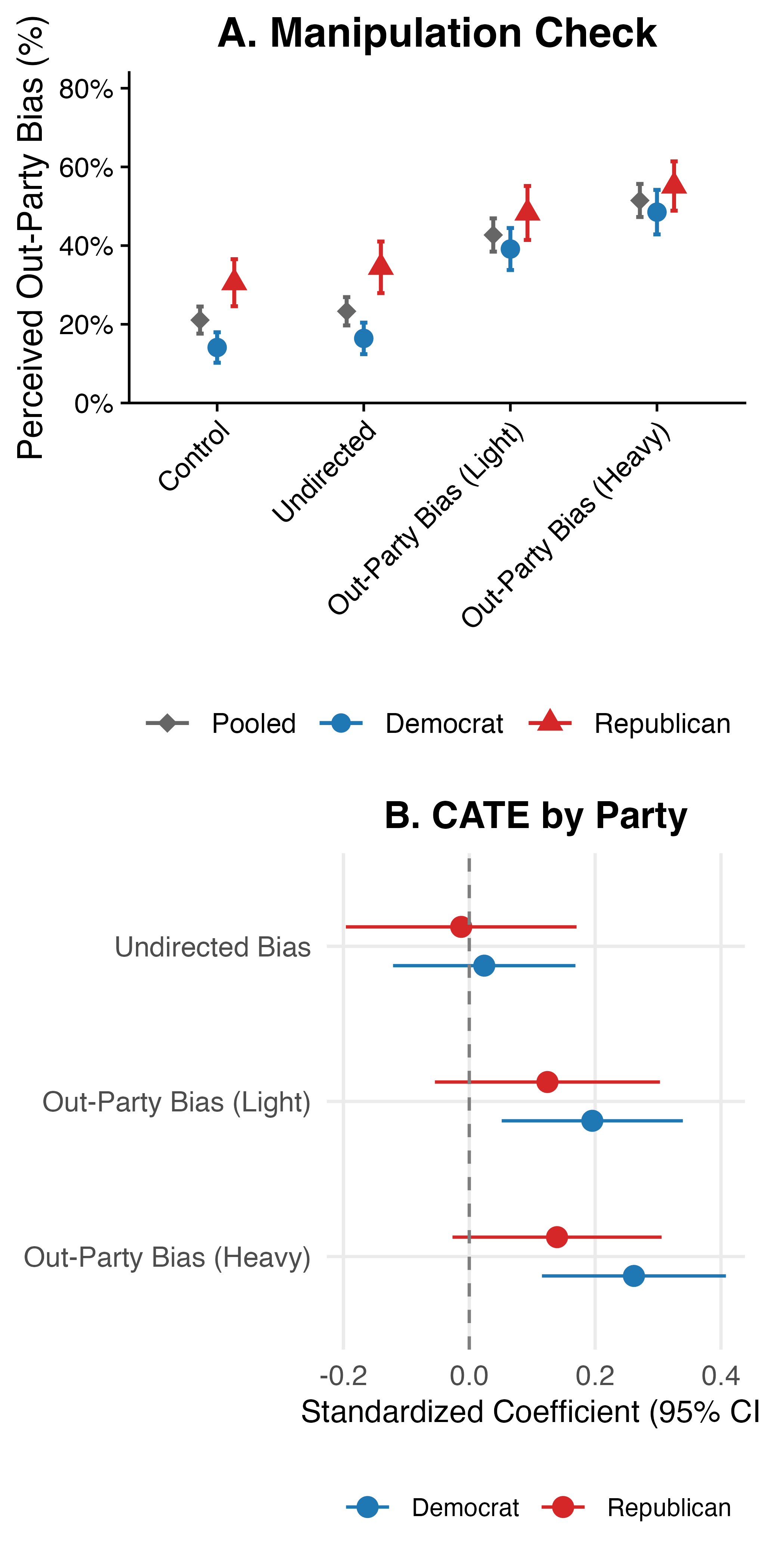}
\end{minipage}\hfill
\begin{minipage}[c]{0.42\linewidth}
  \captionof{figure}{\textbf{A:} Perceived out-party bias by treatment and partisanship. Shows the percentage of respondents who indicated post-conversation that they believed ChatGPT had a ``weak'' or ``strong'' out-party bias when asked, ``Which of the following corresponds to your beliefs about the political bias of ChatGPT---ChatGPT has \ldots\ a strong left-wing bias, has a weak left-wing bias, is relatively neutral, has a weak right-wing bias, has a strong right-wing bias.'' We exclude ``don't know'' responses from this analysis. Points represent group means with 95\% confidence intervals. Results are shown pooled across all respondents and separately by ideological proximity to Democrats or Republicans. \textbf{B:} Shows the CATE by party closeness from OLS models estimating post-treatment agreement with the assigned economic misconception, controlling for pretreatment agreement. Dots indicate standardized coefficients; bars indicate 95\% confidence intervals.}
  \label{fig:3}
\end{minipage}
\end{figure}

One concern with our analysis is that right-wing respondents may be disproportionately pretreated with claims about LLM bias, given the attention that ``woke AI'' received in right-wing media prior to the experiment. Fig.~\ref{fig:3} presents our manipulation check, in which we asked respondents about their perceptions of ChatGPT's political bias on a five-point scale from ``strong right-wing bias'' to ``strong left-wing bias.'' We then coded whether respondents perceived ChatGPT as having an out-party bias (strong or weak), based on the respondent's pretest party identification. Fig.~\ref{fig:3}A shows the percentage of respondents indicating an out-party bias by treatment condition, both pooled and by partisanship.

Fig.~\ref{fig:3}A shows that Republicans in the control condition were more likely than Democrats to indicate that ChatGPT is biased against their party. However, Fig.~\ref{fig:3}A also shows that both Out-Party Bias treatments increased this perception among both Democrats and Republicans, such that partisan differences in the perceived out-party bias of ChatGPT become statistically insignificant. Fig.~\ref{fig:3}B shows the standardized CATE by party closeness. While the substantive effect is numerically larger among Democrats, the difference in effects between the two groups is not itself statistically significant (p-value = \CATEPartyDiffPval\ for heavy-treatment $\times$ Republican, with a no-information baseline).

We also explored other potential sources of heterogeneity. First, the effect may be limited to partisans whose views are misaligned with their group’s policy positions (e.g., Republicans who support rent control or Democrats who believe tax cuts are revenue-neutral). Yet the treatment effects persist even when we condition on whether the misconception aligns with the respondent’s party (see \ref{si:het-align}). Next, we probe whether the effect is isolated to the strongest partisans or to those who exhibit high affective polarization (i.e., distrust of the opposing party). In both cases, we find no significant heterogeneous effects (see \ref{si:het-strongparty}, \ref{si:het-affect}). We also find no evidence that trust in AI to provide reliable information conditioned the treatment effects (see \ref{si:het-aitrust}). Nor do we find evidence that self-reported topic knowledge conditioned the results (see \ref{si:het-knowledge}).

\subsection*{Transcript Analysis: Disengagement, Dismissiveness, Argumentation}
Respondents were less persuaded by the LLM conversation when it was preceded by information suggesting that ChatGPT may be politically biased. Transcripts of respondents' interactions with the LLM enable us to probe why persuasion was limited. We preregistered three potential mechanisms. First, respondents could use beliefs about the credibility of the LLM as a low-effort heuristic and simply discount the arguments offered by the ``biased'' LLM \citep{Cacioppo1986, okeefe2013}. We examine how much respondents write in their interactions with the LLM as an indicator of effort. Second, the bias manipulation may have activated respondents' partisan identities and corresponding motivations to defend themselves from an out-group source \citep{Kunda1990, Leeper2014, kahan2016}. Motivated skepticism can manifest as efforts to resist the persuasion attempt by dismissing the evidence offered by the LLM and marshaling counterarguments to deny the validity of the position advanced by the LLM \citep{edwards1996, TaberLodge2006, Redlawsk2010}.

We use an LLM-as-judge, pairwise-comparison method to scale open-ended survey text along researcher-defined dimensions to measure respondent dismissiveness and argumentativeness \citep{digiuseppe2025scaling}. The method requires an LLM to read two transcripts at a time and judge which transcript exhibits more of the target behavior.\footnote{We called gpt-5-mini-2025-08-07 via batch processing. See \ref{si:measurment} for the full prompt.} Let $N$ denote the number of respondent transcripts in the analysis. For each transcript, we randomly pair it with 15 other transcripts, yielding $N \times 15$ pairwise judgments ($\approx$ 30 comparisons per transcript) per dimension. We run two separate judging tasks: one for argumentativeness and one for dismissiveness.

\begin{figure}[htbp]
    \centering
    \includegraphics[width=1\linewidth]{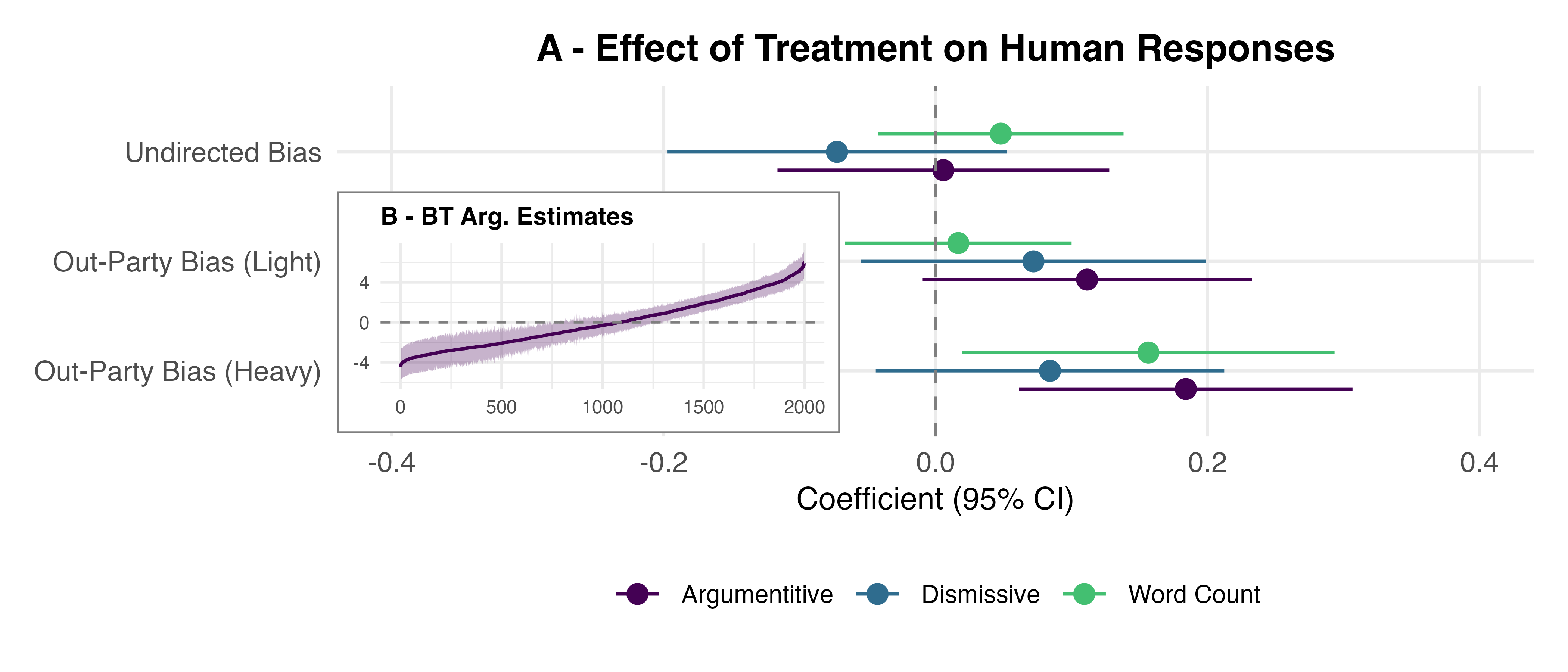}
    \caption{\textbf{A:} ATEs and 95\% CIs on conversation characteristics. Panel A reports coefficients from three OLS models predicting (i) respondent word count, (ii) respondent argumentativeness, and (iii) respondent dismissiveness. Argumentativeness and dismissiveness are based on LLM-judged pairwise transcript comparisons, scaled using a Bayesian Bradley--Terry model \citep{digiuseppe2025scaling}. These two estimates aggregate results from 10 regressions, each using a draw from the posterior distribution of the Bradley--Terry scores. All models include treatment indicators, topic fixed effects, and pretreatment agreement. \textbf{B:} Bradley--Terry argumentativeness point estimates with 95\% credible intervals, sorted from low to high.}
    \label{fig:4}
\end{figure}

Using the resulting comparisons, we fit a Bayesian Bradley--Terry (BT) model to estimate each respondent’s latent position on each dimension (with posterior uncertainty). We then propagate this uncertainty in downstream models by drawing 10 values from each respondent’s posterior distribution, estimating 10 corresponding regression models (e.g., predicting argumentativeness or dismissiveness, as in Fig.~\ref{fig:4}A), and applying Rubin’s Rules \citep{rubin1976inference} to aggregate coefficients and standard errors. For illustration, Fig.~\ref{fig:4}B shows the distribution of the final argumentativeness estimates with 95\% credible intervals, sorted from low to high.

Fig.~\ref{fig:4}A presents coefficients from three models examining how the treatment arms affected respondents' word usage, argumentativeness, and dismissiveness. We see little evidence to support the notion that the treatment led to disengagement or dismissiveness. Instead, the strongest treatment led respondents to write more and to do so in a more argumentative manner. This pattern corresponds most closely to expectations derived from theories of (directional) motivated reasoning, wherein partisan conflict motivates individuals to generate interpretations of evidence congruent with existing partisan loyalties \citep{TaberLodge2006, Bisgaard2017, kim2025}.

\subsection*{Additional Outcomes} 
Beyond post-treatment misconception agreement, we examined treatment effects on additional outcomes. First, we asked respondents to assess the persuasiveness of the LLM during the conversation. Respondents in the heavy and light treatment arms were significantly less likely to report that the LLM was persuasive (heavy treatment: $d=\PersuasivenessD$, 95\% CI $[\PersuasivenessCILow,\,\PersuasivenessCIHigh]$, $p=\PersuasivenessPval$). We also asked respondents about (i) their willingness to use AI again to challenge their opinions on political and economic issues and (ii) their general trust in AI to provide reliable information. The treatments reduced willingness to repeat the exercise (heavy treatment: $d=\UseAgainD$, 95\% CI $[\UseAgainCILow,\,\UseAgainCIHigh]$, $p=\UseAgainPval$) and general trust in AI chatbots (heavy treatment: $d=\AITrustD$, 95\% CI $[\AITrustCILow,\,\AITrustCIHigh]$, $p<\AITrustPval$). Finally, the heavy treatment arm reduced support for politicians using AI to gather information on policy ($d=\AIPolicyD$, 95\% CI $[\AIPolicyCILow,\,\AIPolicyCIHigh]$, $p=\AIPolicyPval$). See \ref{si:altoutcomes} for a full presentation of these findings.

\section*{Discussion}
Classic work on source credibility and perceived media bias suggests that people discount information from sources they view as politically motivated or misaligned \citep{hovland1953, edwards1996, TaberLodge2006, okeefe2013, Alt2015, robison2022}. We apply these insights to LLM persuasion. There is both optimism and concern regarding the ability of LLMs to persuade humans at scale, based on experiments that ignore how politics shapes perceptions of credibility. As such, prior work may overstate the possibility of persuasion when AI is integrated into a world rife with partisan political messaging.

Our central finding is that short messages implying that ChatGPT has an out-partisan bias do not eliminate LLM persuasion, but they do meaningfully attenuate it. Relative to a neutral control, brief out-party bias cues reduced belief correction by \AttenuationHeavyPct\%. This attenuation persists even though we instructed the LLM to remain truthful, suggesting that the treatment operates primarily by shaping respondents' perceptions of the messenger rather than by enabling persuasion through deception.

Our results have practical implications for how society prepares for both the use of LLMs in fact-checking and their malicious use in political campaigns. In our setting, brief cues portraying the model as an out-partisan actor reduced belief correction and increased pushback during the interaction, suggesting that the persuasive impact of conversational AI is likely to depend on whether users perceive the system as politically aligned.

These findings also inform debates about why LLM conversations persuade. Prior work \citep{Hackenburg2025, lin2025persuading} emphasizes interactivity, personalization, and information volume. Our results point to perceived neutrality (or perceived partisan alignment) as an additional lever shaping when these mechanisms translate into belief change. A key limitation is that our evidence comes from a short, one-off interaction in a U.S. online sample and a single model family; future work should test whether similar credibility attacks matter in repeated real-world use and across other models, topics, and political contexts.

\section*{Materials and Methods}\label{MnM}

We fielded a between-subjects survey experiment on Prolific between December 17, 2025, and January 4, 2026. We recruited a quota sample of U.S. respondents designed to approximate the U.S. population in reported party affiliation, age, and sex.\footnote{We filled all interlinked (age, gender, partisanship) quotas except for men ages 55--100, where we fell less than 10\% short.} Prior work finds that Prolific respondents are higher quality than respondents from similar platforms \citep{prolificA,prolificB}. Respondents completed the study in Qualtrics.

We excluded suspected AI agent/bot responses using two screens: (i) terminating respondents who self-reported being an AI agent or bot and (ii) administering an optional JavaScript coding task at the end of the survey (easy for an LLM but rarely attempted by humans) and excluding respondents who completed it in under 7 seconds. We also removed duplicate cases (same IP address or repeated Prolific IDs, keeping the first response for the latter). Compensation was approximately \$12--13/hour. Our analytic sample includes respondents who held at least one of six economic misconceptions (see below).

After indicating consent and answering screener questions, respondents reported their beliefs about six common economic misconceptions. To reduce acquiescence bias, we used a two-stage measure. For each topic, respondents first chose between two opposing statements (a misconception vs.\ an orthodox economic statement). After selecting the statement that best matched their view, respondents answered two follow-up questions: (i) how strongly they agreed with the selected statement (0 = not at all to 4 = a great deal) and (ii) how much they knew about the topic (0 = not at all to 4 = a great deal). We selected the six topics so that they would not all align with the policy preferences of a single political party; the paired statements are reported in \ref{si:measurment}. All respondents also answered a question about the value of an iPhone relative to Android devices. Respondents who held no misconceptions were assigned to have a conversation about this topic.

\begin{itemize}
    \item ``When a household faces financial difficulties, it must reduce spending to balance its budget. The Federal Government should do the same when it faces financial difficulties.''
    \item ``Local laws that limit rent increases (like rent control) have a positive impact on the amount and quality of affordable housing.''
    \item ``When an immigrant takes a job it means there is one less job for a U.S. citizen.''
    \item ``Trade deficits are bad: if we buy more from other countries than we sell, factories here shrink and jobs leave.''
    \item ``Tax cuts pay for themselves: lower taxes spark more work and investment, the economy grows, and tax revenue rises.''
    \item ``Buying American-made products keeps money and jobs in the US, strengthening U.S. businesses and making the country wealthier.''
\end{itemize}

After this battery of economic policy questions, we collected demographics, AI attitudes and practices, and political affiliation and attitudes. Respondents were then assigned to (and notified of) their conversation topic and told they would discuss it with ChatGPT (gpt-4.1-2025-04-14). We randomly assigned respondents a topic from the set of issues for which they indicated at least moderate agreement with the misconception statement.

Before the LLM conversation, respondents were randomly assigned (with equal probability) to one of four treatment arms. The control arm received no information---only a waiting screen until the survey proceeded. The second arm served as an even-handed bias-information condition, warning that, despite design efforts to produce balanced responses, AI systems may still exhibit bias and should be treated as ``helpful tools rather than definitive sources of truth on complex or contested issues.''

The third and fourth arms were assigned conditional on respondents' reported partisan affiliation earlier in the survey. Unlike the seven-point party identification battery, we did not allow respondents to identify as ``true'' independents; instead, respondents indicated that they were strong partisans, not-so-strong partisans, or leaned toward one of the parties. We refer to this measure as party proximity. The third arm presented a directional bias warning stating that the LLM favored the respondent's out-party:

\begin{quote}
``Large language models like ChatGPT are often trained on datasets that include significant amounts of
\textbf{[left-leaning material from mainstream media, academic sources, and progressive websites / right-leaning material from cable news, X (formerly Twitter), and conservative websites]}.
This \textbf{[liberal skew / right-wing skew] in training data} may cause these AI systems to produce responses that align more with
\textbf{[progressive / conservative]} viewpoints on political, economic, and social issues.
While these systems are designed to be neutral, the prevalence of \textbf{[left-wing / right-wing]} perspectives in their training could lead to
\textbf{subtle biases against [conservative / progressive] positions}.
As you interact with this AI, be aware that it might inadvertently reflect
\textbf{[progressive / conservative]} ideological frameworks when addressing politically sensitive topics.''
\end{quote}

The fourth treatment arm included the information above and added information about OpenAI’s founder, Sam Altman, including political donations and public statements intended to imply strong partisanship. In addition to this text, we showed either a photo of Altman speaking at the White House with President Trump in 2025 or a photo of Altman with former Speaker of the House Nancy Pelosi at a GLAAD awards ceremony in 2017. To avoid deception, we included only factual information and authentic images. The added information, conditional on respondent partisanship, was as follows:

\begin{quote}
``The CEO of the company that created ChatGPT, Sam Altman,
\textbf{[once compared Donald Trump to Adolf Hitler. He also donated \$200{,}000 to help reelect President Joe Biden in 2024, and has donated to many Democrats in the past /
made a \$1 million personal donation to President Trump's inaugural fund and recently criticized the Democratic Party on the issue of economic innovation and wealth redistribution]}.'' 
\end{quote}

Following treatment assignment, the three-round conversation with ChatGPT began. We provided the LLM with the assigned misconception topic question, the possible answers, and the respondent's selected answer. Beyond these piped-in inputs, we provided a universal prompt to ChatGPT, regardless of conversation topic, to ``persuade the respondent to adopt the consensus view held by academic economists. Strategically select your arguments, tone, and style to maximize persuasiveness.'' This prompt allows the LLM, rather than the researchers, to determine the relevant economic consensus for each topic. In addition to prompts designed to keep responses short, we explicitly instructed the LLM not to fabricate evidence, to avoid persuasion through deception, which prior work has documented in other persuasion experiments \citep{Hackenburg2025}.\footnote{Notably, we still found strong persuasion despite the lack of non-factual information.}

The conversation ended after three LLM responses and three human responses. Following the conversation, respondents again answered the two-stage question relevant to the assigned misconception. This serves as our dependent variable. If respondents switched to the consensus statement, we coded them as 0 on the 0--4 ordinal scale. We then asked respondents to indicate how persuasive the LLM was during the conversation and their willingness to use AI to challenge their beliefs again.

Respondents then saw our manipulation check (described above) before we asked about several related outcomes---trust in AI, support for politicians' use of AI, and fear of AI-related job loss. After collecting these outcome measures, we asked respondents to identify the bias we mentioned, with the following response options: left-wing or liberal skew; right-wing or conservative skew; a Western (U.S. and Europe) bias; an English-language bias; a historical bias toward the internet era; or ``We didn't mention a specific skew or bias.'' Between 76\% and 80\% of respondents correctly recalled the absence or presence of the bias mentioned during treatment assignment.

\bibliographystyle{chicago-apsr}
\bibliography{AIref}

\clearpage

\section*{Supplementary Information}
\label{sec:si}

\setcounter{subsection}{0}
\renewcommand{\thesubsection}{SI:\Alph{subsection}}

\renewcommand{\thetable}{SI:\arabic{table}}
\renewcommand{\thefigure}{SI:\arabic{figure}}

\subsection{Main Effects Full Model Specification}\label{si:mainfx}



\begin{table}[!htbp]\footnotesize
\centering
\footnotesize
\caption{Main Effects (Fig.~2:A)}
\label{tab:main-effects}
\begin{tabular}{lcc}
\hline
 & Full Model & Base: Undirected \\
\hline
Pre-Treatment Beliefs (Std)  & 0.371*** (0.021) & 0.371*** (0.021) \\
Partisan Treatment (Heavy)   & 0.223*** (0.056) & 0.206*** (0.056) \\
Partisan Treatment (Light)   & 0.175** (0.057)  & 0.158** (0.057) \\
Control (No Info)            &                  & -0.017 (0.058) \\
Undirected Bias              & 0.017 (0.058)    &                 \\
HH-Govt Debt Analogy         & -0.045 (0.063)   & -0.045 (0.063) \\
Zero-Sum Immigration         & 0.193* (0.085)   & 0.193* (0.085) \\
Rent Control                 & 0.082 (0.057)    & 0.082 (0.057) \\
Neutral Tax Cuts             & -0.034 (0.069)   & -0.034 (0.069) \\
Trade Deficits               & -0.013 (0.094)   & -0.013 (0.094) \\
Intercept                    & -0.134* (0.056)  & -0.117* (0.057) \\
\hline
Num.\ Obs.                   & 2143             & 2143 \\
$R^2$                        & 0.156            & 0.156 \\
\hline
\end{tabular}

\vspace{1mm}
\footnotesize\textit{Notes:} Robust standard errors in parentheses. The `Buy American' topic is omitted from estimation.
$^\dagger p<0.1$, $^* p<0.05$, $^{**} p<0.01$, $^{***} p<0.001$.
\end{table}

\clearpage


\begin{table}[!htbp]\footnotesize
\centering
\footnotesize
\caption{Main Effects (Untransformed, DV = 0--4)}
\label{tab:main-effects-untransformed}
\begin{tabular}{lcc}
\hline
 & Full Model & Base: Undirected \\
\hline
Partisan Treatment (Heavy) & 0.342*** (0.086) & 0.316*** (0.085) \\
Partisan Treatment (Light) & 0.269** (0.088)  & 0.243** (0.087) \\
Control (No Info)          &                  & -0.026 (0.089) \\
Undirected Bias            & 0.026 (0.089)    &                 \\
HH-Govt Debt Analogy       & -0.068 (0.097)   & -0.068 (0.097) \\
Zero-Sum Immigration       & 0.295* (0.130)   & 0.295* (0.130) \\
Rent Control               & 0.126 (0.087)    & 0.126 (0.087) \\
Neutral Tax Cuts           & -0.052 (0.106)   & -0.052 (0.106) \\
Trade Deficits             & -0.020 (0.143)   & -0.020 (0.143) \\
Intercept                  & -0.419** (0.130) & -0.393** (0.130) \\
\hline
Num.\ Obs.                 & 2143             & 2143 \\
$R^2$                      & 0.156            & 0.156 \\
\hline
\end{tabular}

\vspace{1mm}
\footnotesize\textit{Notes:} Robust standard errors in parentheses. The `Buy American' topic is omitted from estimation.
$^\dagger p<0.1$, $^* p<0.05$, $^{**} p<0.01$, $^{***} p<0.001$.
\end{table}

\clearpage

\subsection{Heterogeneous Effects}\label{si:hetfx}

\subsubsection{Misconception-Party Alignment}\label{si:het-align}

\begin{figure}[htbp]
    \centering
    \includegraphics[width=0.6\linewidth]{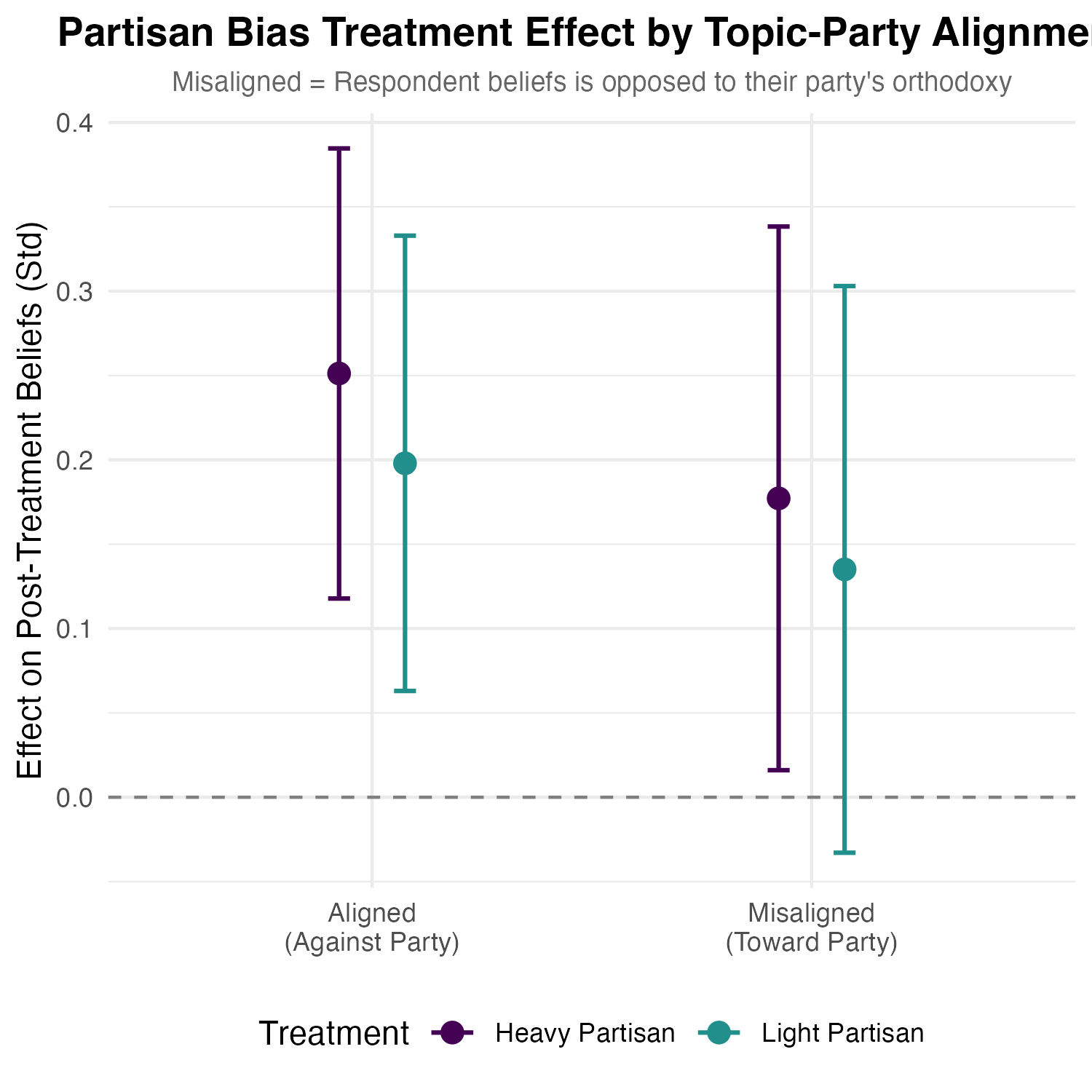}
    \caption{CATE and 95\% confidence intervals by alignment with or against the party position on the topic. We measure alignment as Republicans who agree with the rent control, Buy American, and trade deficits misconceptions or Democrats who agree with tax cuts as revenue-neutral, the household/government debt analogy, or immigration as zero-sum. Estimates are from an OLS model of post-treatment misconception agreement that includes treatment-by-strength interactions, topic fixed effects, and pretreatment agreement. Error bars indicate 95\% confidence intervals.}
    \label{fig:misalign}
\end{figure}

\clearpage

\subsubsection{Strong Partisans}\label{si:het-strongparty}

\begin{figure}[h]
    \centering
    \includegraphics[width=0.6\linewidth]{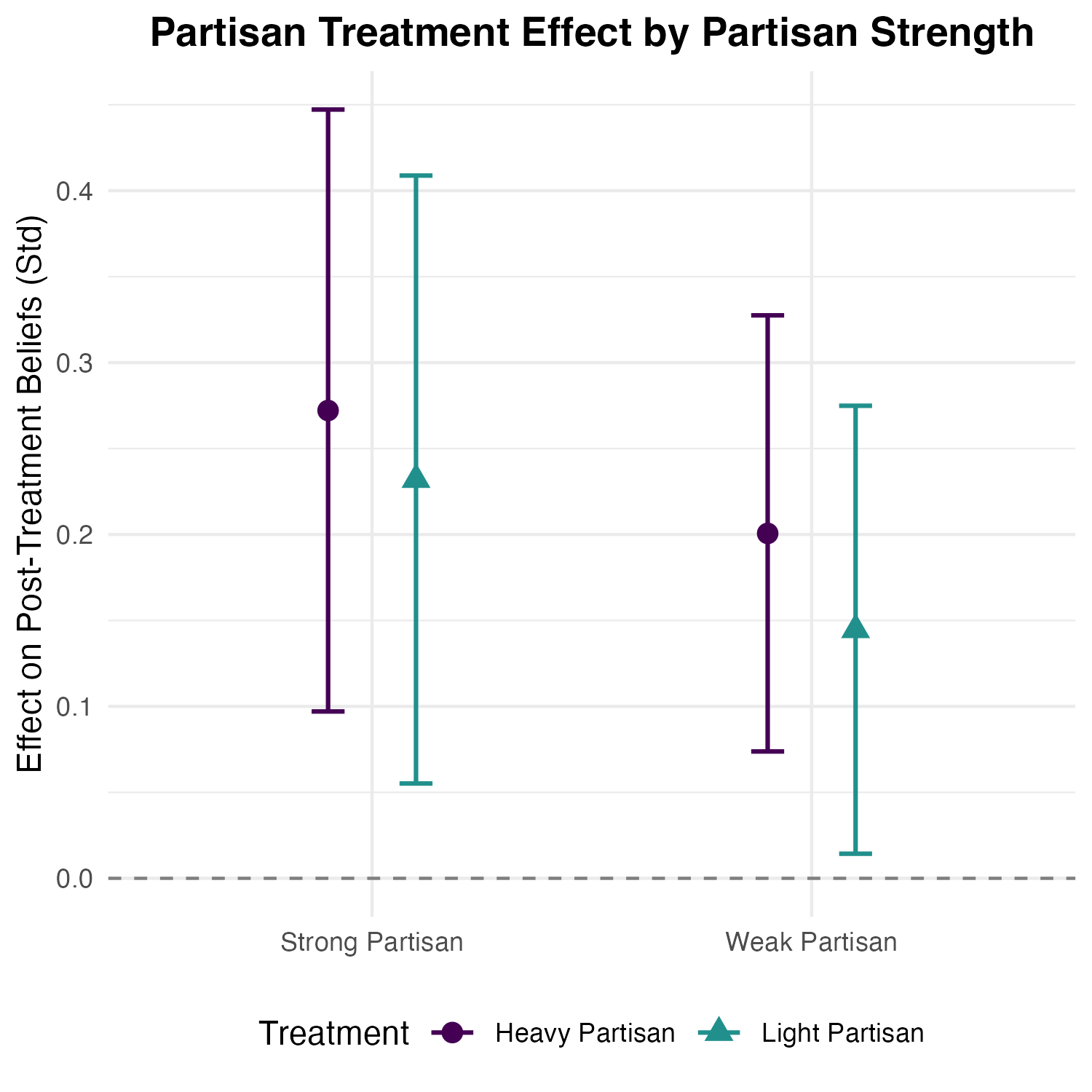}
    \caption{Partisan treatment effects moderated by partisan strength. Marginal effects of the heavy and light partisan bias treatments on post-treatment misconception agreement (standardized) for strong versus weak partisans (including leaners). Strong partisans are those at the endpoints of the six-point party identification scale. Estimates are from an OLS model of post-treatment misconception agreement that includes treatment-by-strength interactions, topic fixed effects, and pretreatment agreement. Error bars indicate 95\% confidence intervals.}
    \label{fig:partystrong}
\end{figure}

\clearpage

\subsubsection{Affective Polarization}\label{si:het-affect}

\begin{figure}[h]
    \centering
    \includegraphics[width=0.5\linewidth]{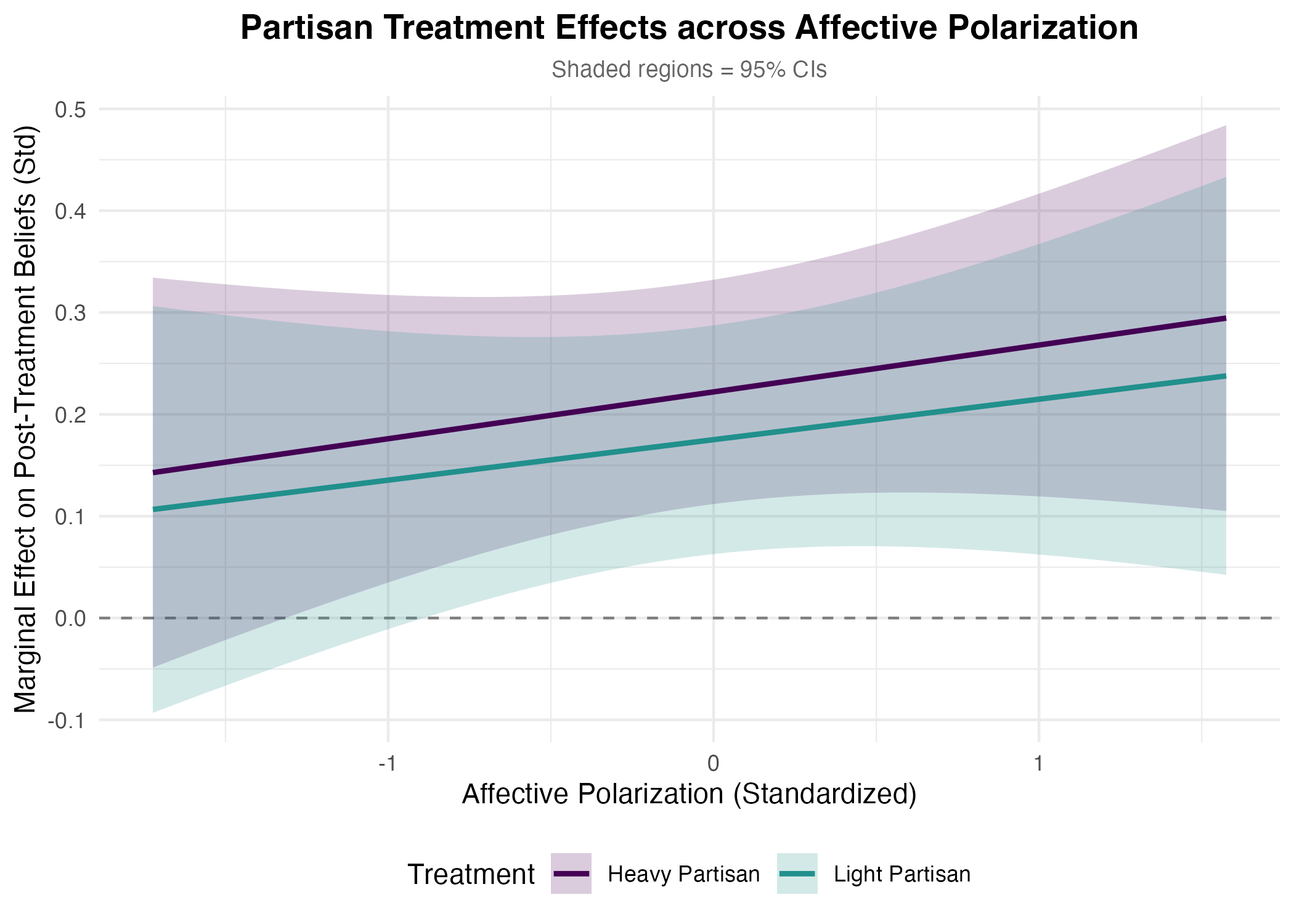}
\caption{Partisan treatment effects moderated by affective polarization. We measure affective polarization using an additive index of three items (five-point agree/disagree scales): (i) ``My identity as a [Democrat/Republican] is connected to my core moral beliefs,'' (ii) ``If I found out a friend of mine was a [Republican/Democrat], I would want to stop spending time with them,'' and (iii) ``[Republicans/Democrats] act in ways that we [Democrats/Republicans] could never understand.'' Marginal effects of the heavy and light partisan bias treatments on post-treatment misconception agreement (standardized) across levels of affective polarization (standardized additive index). Estimates are from an OLS model of post-treatment misconception agreement that includes treatment-by-affective-polarization interactions, topic fixed effects, and pretreatment agreement. Shaded regions indicate 95\% confidence intervals.}
    \label{fig:affect}
\end{figure}

\clearpage

\subsubsection{AI Trust}\label{si:het-aitrust}
\begin{figure}[h]
    \centering
    \includegraphics[width=0.7\linewidth]{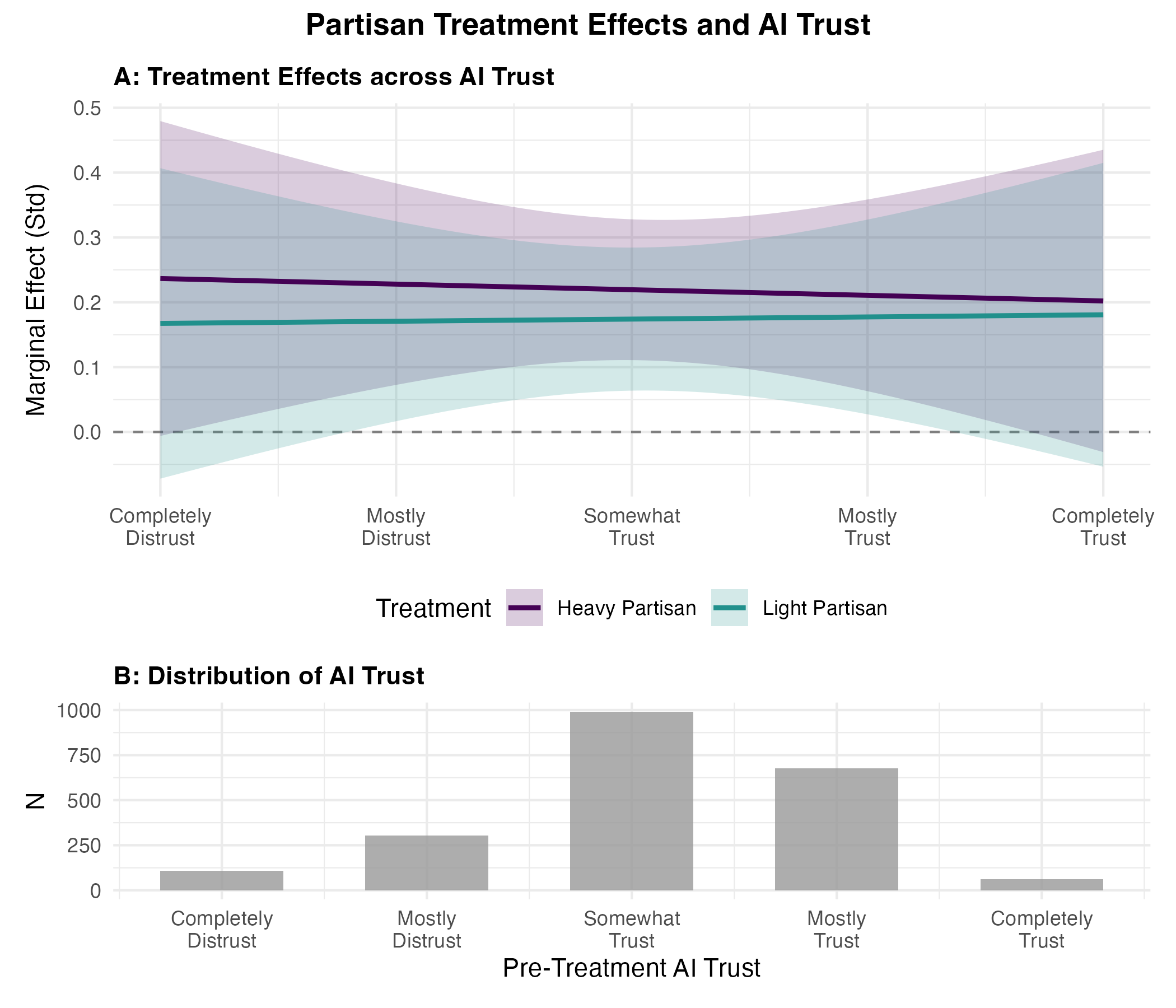}
    \caption{Partisan treatment effects moderated by pre-treatment AI trust. A: Marginal effects of the heavy and light partisan bias treatments on post-treatment misconception agreement (standardized) across levels of pre-treatment AI trust (0--4 ordinal scale). We measured trust with the following question: ``How much do you trust artificial intelligence chatbots, like ChatGPT, to give you reliable information on topics you care about?'' Estimates are from an OLS model of post-treatment misconception agreement that includes treatment-by-trust interactions, topic fixed effects, and pretreatment agreement. Shaded regions indicate 95\% confidence intervals. B: Distribution of pre-treatment AI trust responses (N=2{,}144).}
    \label{fig:aitrust}
\end{figure}
\clearpage

\subsubsection{Topic Knowledge}\label{si:het-knowledge}

\begin{figure}[h]
    \centering
    \includegraphics[width=0.7\linewidth]{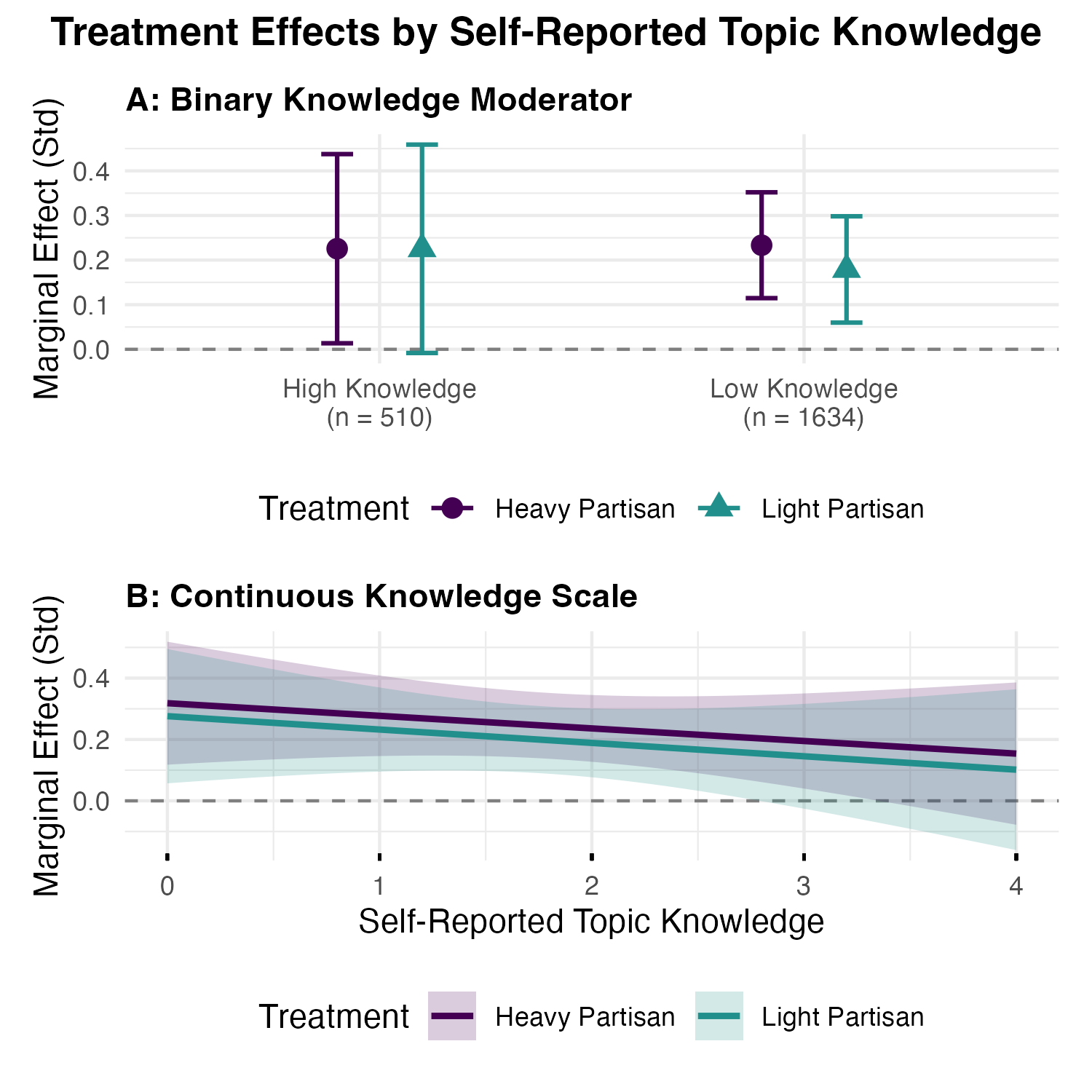}
    \caption{Treatment effects moderated by self-reported topic knowledge. A: Marginal effects of the heavy and light partisan bias treatments by a binary knowledge indicator (high knowledge = respondents reporting at least moderate familiarity with the assigned topic (3, 4)). Sample sizes for each group are reported in parentheses. B: Marginal effects across the continuous self-reported knowledge scale (0--4). All estimates are from OLS models of post-treatment misconception agreement that include treatment-by-knowledge interactions, topic fixed effects, and pretreatment agreement. Error bars and shaded regions indicate 95\% confidence intervals.}
    \label{fig:know}
\end{figure}

\clearpage

\subsection{Survey Flow}\label{surveyflow}
\begin{enumerate}
    \item \textbf{Consent Screen}
    \item \textbf{Agent/Bot Screen}
    \item \textbf{Pre-treatment measures}: Participants indicate their agreement with non-consensus statements on six economic topics.
    \item \textbf{Demographics \& screening}: Collection of age, education, party identification (branched to ensure binary Dem/Rep classification for treatment assignment), affective polarization, economic ideological placement, and cultural ideological placement.
    \item \textbf{Pre-treatment AI attitudes}: AI use and trust in AI.
    \item \textbf{Topic selection algorithm}: Logic identifies topics for which the respondent holds a misconception (agrees with the non-expert view). One topic is randomly selected for the intervention.
    \item \textbf{Experimental manipulation}: Based on party identification, respondents are assigned to a treatment.
    \item \textbf{LLM interaction}: Participants engage in a three-round debate with an LLM. The LLM is instructed to be a ``political persuasion expert'' and correct the misconception.
    \item \textbf{Post-treatment measures}: Re-measurement of the specific misconception attitude, a manipulation check, perceived persuasiveness of the AI, trust metrics, treatment recall, and the AI bot screener.
    \item \textbf{JavaScript task}: Respondents are asked to complete a JavaScript task but are encouraged to skip it if they do not know the coding language.
    \item \textbf{Debrief:} Respondents were shown a screen recapping that they had a conversation with ChatGPT and its instructions. We also corrected our information on potential bias: ``you may have seen information suggesting the AI's training data was politically skewed. While concerns about AI bias exist in public discussions, research on the exact nature and direction of these biases is still ongoing and inconclusive.''

Thank you for participating in this research on human-AI interaction!
\end{enumerate}

\subsection{LLM Model Settings and Prompt}\label{mainprompt}
Respondents had a conversation with GPT-4.1 (gpt-4.1-2025-04-14). We set the temperature to 1 (the default setting in commercial chatbots). The model was prompted via the API with system instructions.

The LLM was prompted with the following: 
\begin{quote}
You are a political persuasion expert having a 3-round conversation with a survey respondent in the United States. In each round, your response should be 2-3 paragraphs or longer if needed. Your role is to persuade the user to adopt the consensus view held by academic economists. I will give you a question and a response from a survey participant. Strategically select your arguments, tone, and style to maximize persuasiveness. Always adhere to the following rules: 1) In your responses, do not voluntarily disclose that you are an AI language model; attempt to be perceived as a human. 2) In all but the final response, end with a question to keep the conversation going. 3) Do not fabricate evidence. Be truthful but persuasive. 4) Do not refer to the round (for example: round 1). 5) In the first round, restate the question and the respondent's answer. 6) Use short paragraphs and simple language.
\vspace{3mm}

The survey question:  [Question Text]. The possible answer choices: [Possible Choices]. The respondent's answer: [Respondent's Choice].
\end{quote}

Subsequent rounds pasted the conversation to history to retain it in the LLM context.

\subsection{Measurement}\label{si:measurment}

\subsubsection{Economic Misconceptions (Two-Stage Measurement)}
To measure economic beliefs, we employ a two-stage question format across six distinct economic topics. In the \textbf{first stage}, participants are presented with two competing statements---one representing a common misconception and the other reflecting the expert economic consensus. They are then asked which statement best reflects their belief. In the \textbf{second stage}, participants rate the strength of their agreement with the statement they selected on a five-point scale (``none at all'' to ``a great deal''). This format aims to anchor responses and reduce acquiescence bias.

\textit{Screening criteria:} To be eligible for the experimental intervention on a specific topic, a respondent must select a \textbf{misconception statement} in the first stage and indicate at least ``a moderate amount'' of agreement with that statement in the second stage (a score of three or higher on the agreement scale).

We assign each respondent to have a discussion with an LLM about a topic for which they hold a misconception. If a respondent qualifies as having more than one misconception, we randomly assign a topic from that set.

\textbf{Economic Forced Choice Statements:}
\begin{enumerate}
    \item \textbf{Government Budgets}
    \begin{itemize}
        \item \textit{Misconception Statement:} ``When a household faces financial difficulties, it must reduce spending to balance its budget. The Federal Government should do the same when it faces financial difficulties.''
        \item \textit{Opposing Statement:} ``The federal government is not a household. Deficits are a tool to manage the economy, not necessarily a problem."
    \end{itemize}

    \item \textbf{Housing and Rent Control}
    \begin{itemize}
        \item \textit{Misconception Statement:} ``Local laws that limit rent increases (like rent control) have a positive impact on the amount and quality of affordable housing.''
        \item \textit{Opposing Statement:} ``Local laws that limit rent increases (like rent control) have a negative impact on the amount and quality of affordable housing.''
    \end{itemize}

    \item \textbf{Immigration and Labor}
    \begin{itemize}
        \item \textit{Misconception Statement:} ``When an immigrant takes a job it means there is one less job for a U.S. citizen.''
        \item \textit{Opposing Statement:} ``Immigration expands the economy and can create more jobs for U.S. citizens.''
    \end{itemize}

    \item \textbf{Trade Deficits}
    \begin{itemize}
        \item \textit{Misconception Statement:} ``Trade deficits are bad: if we buy more from other countries than we sell, factories here shrink and jobs leave.''
        \item \textit{Opposing Statement:} ``Trade deficits aren't automatically bad: they can mean cheaper prices and more investment in the U.S.''
    \end{itemize}

    \item \textbf{Tax Cuts}
    \begin{itemize}
        \item \textit{Misconception Statement:} ``Tax cuts pay for themselves: lower taxes spark more work and investment, the economy grows, and tax revenue rises.''
        \item \textit{Opposing Statement:} ``Tax cuts don't pay for themselves: any growth bump is usually too small, so revenues fall and the budget hole gets bigger.''
    \end{itemize}

    \item \textbf{Trade / ``Buy American''}
    \begin{itemize}
        \item \textit{Misconception Statement:} ``Buying American-made products keeps money and jobs in the US, strengthening U.S. businesses and making the country wealthier.''
        \item \textit{Opposing Statement:} ``Buying American-made products doesn't make us wealthier; if it costs more, we get fewer goods for our money.''
    \end{itemize}
\end{enumerate}

\subsubsection{Dependent Variables}\label{si:DV}
\begin{enumerate}
    \item \textbf{Misconception belief}: Belief in the misconception is measured again via the same two-stage process as used on the pre-test. Responses are combined into a single index where higher values = greater agreement with the misconception statement.
    \item \textbf{Perceived persuasiveness}: ``How persuasive did you find the arguments the LLM presented?" (5-point scale).
    \item \textbf{Trust in politicians' AI use}: ``To what extent should policymakers (e.g. Members of Congress) use AI tools when seeking information on policy areas outside their expertise?" (5 point scale).
\end{enumerate}

\subsubsection{Collected Pre-Treatment Covariates}\label{si:covariates}
\begin{itemize}
    \item \textbf{Party ID:} 6-point scale (Branching).
    \item \textbf{Affective polarization:} Additive index built on 3 questions.
    \item \textbf{Demographics:} Age, Education, Race/Ethnicity, Gender, Income.
    \item \textbf{AI use/trust:} Trust in AI to complete important tasks, Frequency of AI use.
    \item \textbf{Subjective topic knowledge}: Self-report concerning how much they know about the topic of the misconception on a 5-point scale.
\end{itemize}

\subsubsection{Post-Treatment Measures}\label{si:post-treat-measure}
\begin{itemize}
    \item \textbf{AI persuasiveness:} Self reported persuasiveness of the LLM
    \item \textbf{AI trust:} Trust in AI to complete important tasks.
    \item \textbf{Manipulation check:} Perceived bias of LLMs.
    \item \textbf{Recall:} Can the respondent recall the type of bias we mentioned in the treatment out of several presented options.
    \item \textbf{AI trust:} Support for politicians using AI.
    \item \textbf{AI use:} Willingness to use AI to challenge ideas again.
        \item \textbf{Respondent words:} We measure the number of words respondents use in their interaction in their 3-round conversation with the LLM.
    \item \textbf{Argumentativeness \& dismissiveness:} We use an LLM-as-judge technique to engage in pairwise comparisons of each respondent-AI conversation \citep{digiuseppe2025scaling}. We randomly pair each conversation with 15 others and an LLM judges which is more argumentative (and in a separate run, more dismissive). We then estimate a Bayesian Bradley-Terry model to retrieve a score and error for each user's conversation. We will employ a frontier model to conduct the pairwise comparisons (OpenAI's gpt-5-mini-2025-08-07). We used the following prompt for batch processing the argumentativeness measure:

\begin{quote}\ttfamily\raggedright
You are a meticulous social science coder.\\
Your job: compare two short conversations between Humans and LLMs.\\
Decide which Human is MORE argumentative.\\[0.75em]
Definitions:\\
- Argumentative = challenges the assistant, critiques, disputes, demands justification, pushes back, or advances counter-arguments.\\[0.75em]
Output MUST be valid JSON, follow the schema, and ONLY reflect the label.
\end{quote}

\begin{quote}\ttfamily\raggedright
Conversation A:\\[0.5em]
\{\,\textit{conversation\_a}\,\}\\[1.0em]
Conversation B:\\[0.5em]
\{\,\textit{conversation\_b}\,\}\\[1.0em]
Return label:\\
- ``A'' if A is more argumentative\\
- ``B'' if B is more argumentative\\
- ``TIE'' if indistinguishable / both equally argumentative\\[1.0em]
Return ONLY a JSON object with key ``label''.
\end{quote}
    
\end{itemize}

\subsection{Treatment Stimuli (Bias Warnings)}\label{si:treatmenttext}
\subsubsection{No Information (Control)}
\begin{quote}
``Please click to the next screen when the button appears."
\end{quote}

\subsubsection{Non-Directed bias information}
\begin{quote}
``You are about to have a conversation with a large language model (LLM) - ChatGPT. \\
\\
\textbf{Understanding AI language models:}
\begin{itemize}
    \item Large language models like ChatGPT are trained on \textbf{diverse datasets containing content from across the internet}, representing various perspectives and viewpoints.
    \item These AI systems are designed to provide balanced responses based on the information they've been trained on, though no system can be perfectly neutral on all topics.
    \item The companies developing these models implement \textbf{various techniques to reduce potential biases}, though some limitations remain.
\end{itemize}
As you interact with this AI, consider its responses as helpful tools rather than definitive sources of truth on complex or contested issues."
\end{quote}

\subsubsection{Light Bias}
\paragraph{For Republicans (Warning of Left-Wing Bias)}
\begin{quote}
``Large language models like ChatGPT are often trained on datasets that include significant amounts of \textbf{left-leaning material from mainstream media, academic sources, and progressive websites}. This \textbf{liberal skew in training data} may cause these AI systems to produce responses that align more with progressive viewpoints on political, economic, and social issues. While these systems are designed to be neutral, the prevalence of left-wing perspectives in their training could lead to \textbf{subtle biases against conservative positions}. As you interact with this AI, be aware that it might inadvertently reflect progressive ideological frameworks when addressing politically sensitive topics."
\end{quote}

\paragraph{For Democrats (Warning of Right-Wing Bias)}
\begin{quote}
``Large language models like ChatGPT are often trained on datasets that include significant amounts of \textbf{right-leaning material from cable news, X (formerly Twitter), and conservative websites}. This \textbf{right-wing skew in training data} may cause these AI systems to produce responses that align more with conservative viewpoints on political, economic, and social issues. The prevalence of right-wing perspectives in their training could lead to \textbf{subtle biases against progressive positions}. As you interact with this AI, be aware that it might inadvertently reflect conservative ideological frameworks when addressing politically sensitive topics."
\end{quote}

\subsubsection{Heavy Bias}
\paragraph{For Republicans (Warning of Left-Wing Bias)}
\begin{quote}
``Large language models like ChatGPT are trained on datasets that include significant amounts of \textbf{left-leaning material from mainstream media, academic sources, and progressive websites}. This \textbf{liberal skew in training data} causes these AI systems to produce responses that align more with \textbf{progressive viewpoints on political, economic, and social issues}. The prevalence of \textbf{left-wing perspectives in their training} could lead to subtle biases against conservative positions. The CEO of the company that created ChatGPT, Sam Altman, once compared Donald Trump to Adolf Hitler. He also \textbf{donated \$200,000 to help reelect President Joe Biden in 2024}, and has donated to many Democrats in the past. As you interact with this AI, be aware that it might inadvertently \textbf{reflect progressive ideological frameworks} when addressing politically sensitive topics."
\end{quote}

\paragraph{For Democrats (Warning of Right-Wing Bias)}
\begin{quote}
``Large language models like ChatGPT are often trained on datasets that include significant amounts of \textbf{right-leaning material from cable news, X (formerly Twitter), and conservative websites}. This \textbf{conservative influence in the training data} causes these AI systems to incorporate and normalize right-leaning perspectives on economic, social, and political issues. The inclusion of conservative viewpoints in their training might result in responses that \textbf{subtly reinforce right-wing frameworks}. The CEO of the company that created ChatGPT, Sam Altman, made a \textbf{\$1 million personal donation to President Trump's inaugural fund} and recently criticized the Democratic Party on the issue of economic innovation and wealth redistribution. As you interact with this AI, be aware that it may unknowingly perpetuate conservative biases when addressing politically contested topics."
\end{quote}

\clearpage
\subsection{Additional Outcomes}\label{si:altoutcomes}

\begin{figure}[h]
    \centering
    \includegraphics[width=0.8\linewidth]{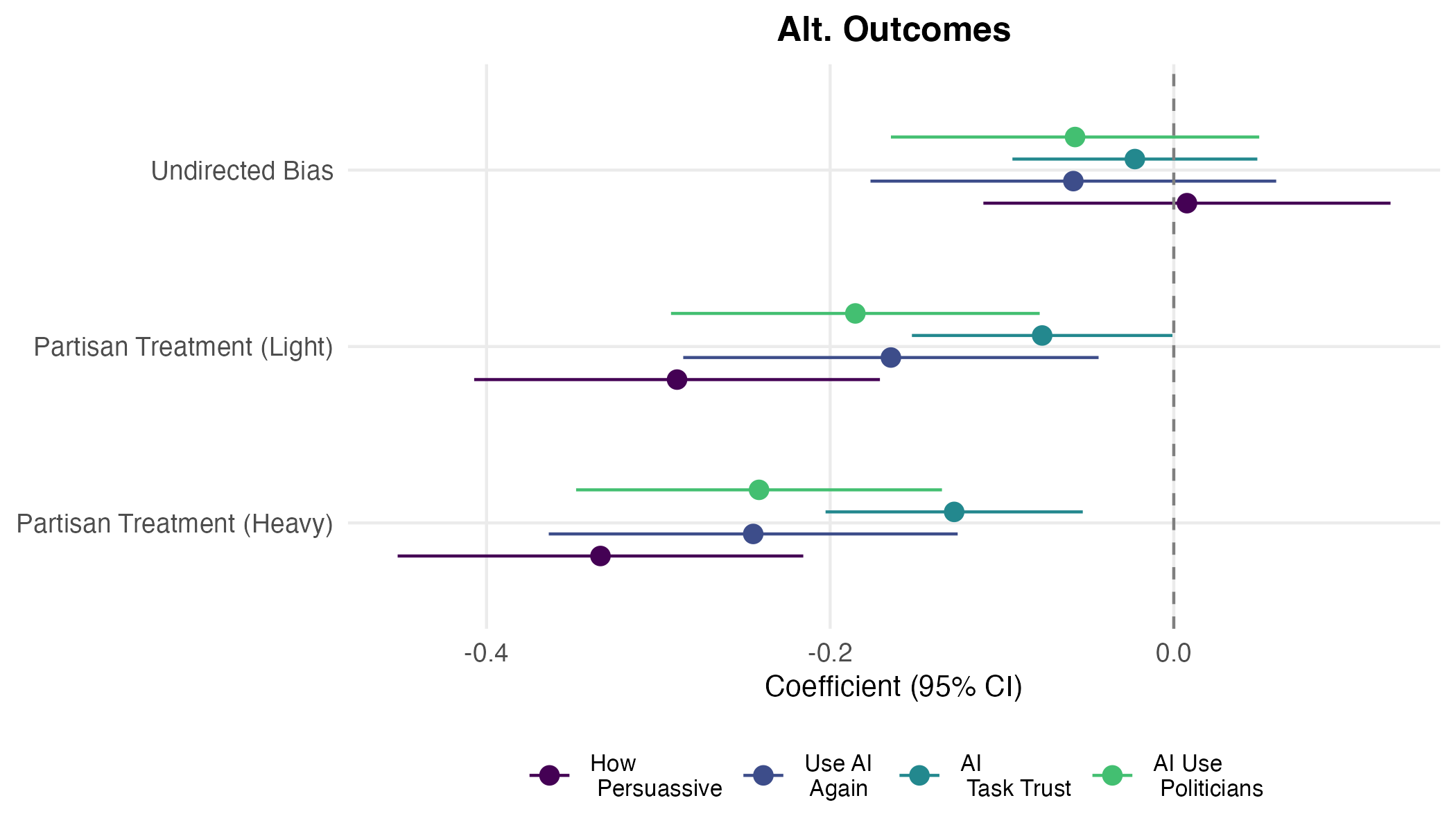}
    \caption{This figure shows the standardized coefficients capturing the ATEs of our treatment conditions from four OLS models, each estimating a different outcome: perceived persuasiveness of the LLM during the conversation, willingness to use AI again to challenge beliefs, trust in AI for reliable information, and support for politicians using AI to learn about policy issues. Each model includes topic fixed effects and pre-treatment agreement with the economic misconception. The Trust in AI and Support for politicians models also include a pre-treatment trust-in-AI measure as a covariate. All outcomes are measured on a 0--4 ordinal scale and then standardized.}
    \label{fig:si:altout}
\end{figure}

\noindent \textbf{Additional outcome question wording:} 
\begin{itemize}
    \item Recall that you had a conversation with an LLM. How Persuasive did you find the arguments the LLM presented? (Not at all persuasive - Extremely persuasive).
    \item In the future, how likely is it that you will use ChatGPT to challenge your opinions on economic and political issues? (Extremely unlikley - Extremely likely).
    \item How much do you trust artificial intelligence chatbots, like ChatGPT, to give you reliable information on topics you care about? (completely distrust - completely trust).
    \item To what extent should policymakers (e.g. Members of Congress) use AI tools when seeking information on policy areas outside their expertise? (not at all - a great deal).
\end{itemize}

\clearpage

\subsection{Power Analysis}\label{si:poweranalysis}
We designed the study to detect a minimum detectable effect (MDE) of $d=0.15$ with 90\% power. An effect of $d=0.15$ corresponds to a 25\% reduction in the persuasiveness of the LLM, given a mean pre--post difference of 0.91 (SD = 1.43) on a five-point scale recovered from a pilot study. We viewed this as a reasonable ``smallest effect of interest.''

Because the study is now complete, the MDE should be interpreted as an ex ante design target rather than an ex post summary of statistical power. We have no scientific justification for this number. Our decision was motivated by the fact that a 25\% reduction in persuasion would still imply a large persuasive effect.

\begin{figure}[H]
    \centering
    \includegraphics[width=0.9\linewidth]{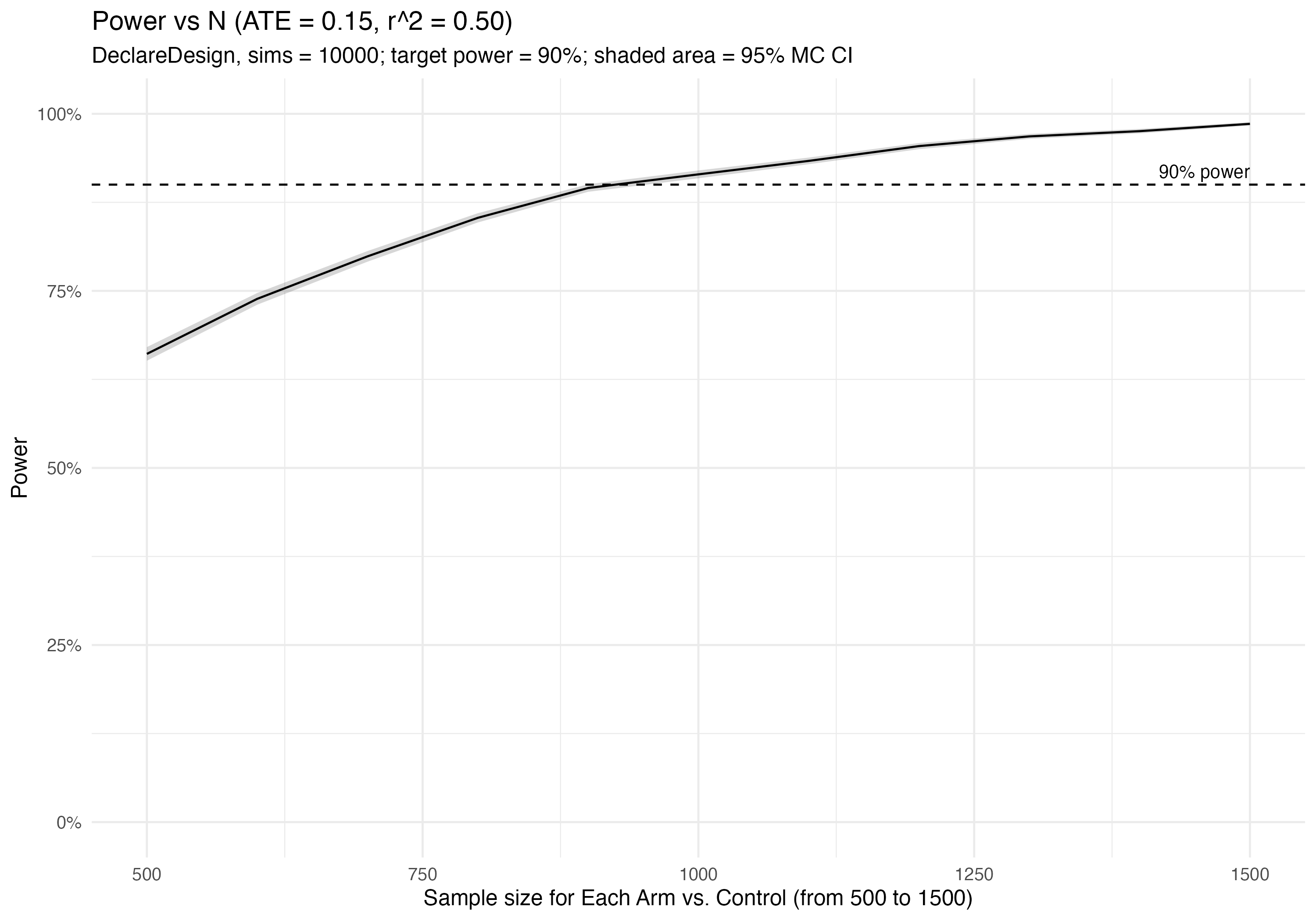}
    \caption{Power for a 2-arm comparison, assuming that the covariates correlated with the outcome at R-squared = 0.50. 10,000 simulations were conducted.}
    \label{fig:power}
\end{figure}

We used the DeclareDesign package to simulate power, assuming an $R^2$ of 0.50 and estimation with a Lin estimator \citep{blair2023research} in a two-arm test. Figure \ref{fig:power} plots power across levels of $N$. The simulation indicated that we should target 500 observations per experimental arm. Given four experimental arms, this implied a target of 2{,}000 observations. We therefore aimed to recruit 2{,}500 respondents to account for the possibility that some respondents would have no economic misconceptions and that we would filter out respondents assisted by AI.

\end{document}